\documentclass{article}

\usepackage[final, nonatbib]{nips_2016}

\usepackage[utf8]{inputenc} 
\usepackage[T1]{fontenc}    
\usepackage{hyperref}       
\usepackage{url}            
\usepackage{booktabs}       
\usepackage{amsfonts}       
\usepackage{amsmath}        
\usepackage{mathrsfs}       
\usepackage{nicefrac}       
\usepackage{microtype}      
\usepackage{bm}             
\usepackage{graphicx}
\usepackage{xcolor}
\usepackage{subcaption}
\hypersetup{
    colorlinks,
    linkcolor={red!50!black},
    citecolor={blue!50!black},
    urlcolor={blue!80!black}
}

\graphicspath{ {./} }

\newcommand{\argmin}{\mathop{\rm arg\,min}\limits} 
\newcommand{\argmax}{\mathop{\rm arg\,max}\limits} 

\title{Emotion Recognition From Speech With Recurrent Neural Networks}

\author{
  Vladimir Chernykh\\
  MIPT, Skoltech\\
  Moscow\\
  \texttt{vladimir.chernykh@phystech.edu} \\
  \And
  Pavel Prikhodko \\
  Skoltech\\
  Moscow \\
  \texttt{p.prikhodko@skoltech.ru} \\
}

\begin{document}

\maketitle


\begin{abstract}
In this paper the task of emotion recognition from speech is considered. Proposed approach uses deep recurrent neural network trained on a sequence of acoustic features calculated over small speech intervals. At the same time special probabilistic-nature CTC loss function allows to consider long utterances containing both emotional and neutral parts. The effectiveness of such an approach is shown in two ways. Firstly, the comparison with recent advances in this field is carried out. Secondly, human performance on the same task is measured. Both criteria show the high quality of the proposed method.
\end{abstract}


\section{Introduction} \label{sec:introduction}

Nowadays machines can successfully recognize human speech. Automatic speech recognition (ASR) services can be found everywhere. Voice input interfaces are used for many applications from navigation system in mobile phones to Internet-of-Things devices. A lot of personal assistants like Apple Siri \cite{apple18siri}, Amazon Alexa \cite{amazon18alexa}, Yandex Alisa \cite{yandex18alisa}, or Google Duplex \cite{google18duplex} were released recently and are already the inalienable part of the life.

Nevertheless this field is still rapidly emerging. Last year Google has released its Cloud API for speech recognition \cite{google18speechapi}. In the last Windows 10 one can find Cortana voice interface \cite{microsoft17cortana} integrated. Small startups all over the world as well as IT giants like Google, Microsoft and Baidu are actively doing research in this area. 

Market size of both hardware and software for speech recognition has reached 55 billion dollars in 2016 and it continues to grow approximately 11\% a year \cite{gvr2018voicemarket}.

Therefore authors believe that this field is perspective and is worth to put an attention at.


\subsection{Problem} \label{ssec:problem}

Virtually all the ASR algorithms and services are simply transcribing audio recordings into written words. But that is only the first level of speech understanding.

During the conversation humans receive lots of meta-information apart from text. Examples might be the person who is speaking, his intonation and emotion, loudness, shades etc. These factors might considerably influence the true intended meaning of a phrase. Even turn it into opposite - that is what we call sarcasm or irony. Humans take all these elements into consideration while processing the phrase in the brain and only after that the final meaning is formed.

Accounting for these factors in purely retrieval systems, e.g. search engines, may be superfluous. But it becomes crucial in more human-involved systems like voice assistants, where the close communication with human is needed. To be able to detect the meaning of the spoken message correctly one needs to account not only for the semantics but also for the discussed type meta-information. Thus to build a more complete human-computer interaction system it is necessary to extract these features out of the audio signal.

This paper addresses only one of the questions arisen above: how to correctly recognize the emotional background of the voice? The main goal of the work is to answer this question. The main obstacles that complicate the solution are:
\begin{itemize}
    \item Emotions are subjective. 
    
    They are complex psychological and social phenomena. People understand emotion differently. Thus there are many difficulties in defining the notion of emotion \cite{devillers2005challenges}.
    
    Altrov et al. in \cite{altrov2015humanemotions} collected the corpus of Estonian speech with 4 emotions included: joy, anger, sadness, neutral. Then they asked people of different nationalities to evaluate it. Estonians, Latvians, Italians, Finns, Swedes, Danes, Norwegians and Russians took part in the experiment. Almost all nationalities are close to Estonians both geographically and culturally. Nevertheless Estonians perform much better than any other nationality showing about 69\% mean class accuracy. All other people perform 10-15\% worse and the only emotion that they recognize relatively well was sadness.
    
    Work of Altrov et al. \cite{altrov2015humanemotions} showed that there is significant intercultural differences in emotions understanding. But even inside one culture this understanding may vary greatly.
    
    \item Assignment of the emotions to the audio recording.
    
    It is not obvious how one should assign emotional labels to the long audio recording or even continuous flow of speech. Should it be one emotion per whole recording or per one utterance? If one chooses utterance-based solution then how the split should be done? Is it possible for the utterance to have multiple emotions? These and few other questions put the methodology in the forefront.
    
    \item Complexity and cost of database collection.
    
    Databases for usual speech recognition task are relatively easy to collect: one can take dialogues from the films, Youtube blogs, news, etc. and annotate them. Almost the only requirement is the high quality of the audio recording.
    
    When it comes to the emotions there is a huge problem with all of these sources. Emotions in them are dramatically biased. In news most of the speech is neutral. In films set of emotions depends on the genre but the distribution is almost always biased towards the one prevailing emotion.
    
    Another way is to collect the database artificially. The following big problem arises here: how to record a predefined emotion in a natural way? Douglas-Cowie et al. suggest to use professional actors \cite{douglascowie2003databasegeneration}. Actors are given either with the topics and asked to improvise on this topic or with the scripted material which they should read. At the time of reading actors are to show the predefined emotion. Busso et al. give the overview and the comparison of these two approaches in their paper \cite{busso2008scriptedorimprovised}.
    
    The set of emotions to use is another important question. There should enough emotions to cover all the basic human reactions but not too many to be able to play and assess them reliably. Picard et al. describe the how and why the emotions should be chosen in their work \cite{picard1995commonemo}. They suggest to use at least 5 basic emotions: happiness, anger, sadness, neutral, frustration.
    
    The other side of this coin is how the emotions should be measured and evaluated. Cowie et al. give their view to this problem in their paper \cite{cowie2003emotionsdescription}. Authors propose to use 3D Valence-Arousal-Dominance ordinal space as well as categorical labels for the evaluation of the utterances. Moreover, many assessors are needed for one utterance to be able to evaluate it consistently.
    
    Altogether, these peculiarities make the collection of the database very complicated, time-consuming and expensive task.
    
    One of the good methodology and collection examples is IEMOCAP database presented by Busso et al. in \cite{busso2008iemocap}. IEMOCAP is used in this work and will be described in more details later.
\end{itemize}

Some of these questions are resolved by authors of this paper, others are tackled by the authors of database used, third are inherent to the problem and can not be avoided.


\subsection{Related works} \label{ssec:related}

The problem described in section \ref{ssec:problem} has previously been considered by few works.

Majority of the works state the emotion recognition task as a classification problem where one utterance has exactly one label.

Before the deep learning era people have come with many different methods which mostly extract complex low-level handcrafted features out of the initial audio recording of the utterance and then apply conventional classification algorithms. One of the approaches is to use generative models like Hidden Markov Models or Gaussian Mixture Model to learn the underlying probability distribution of the features and then to train a Bayessian classifier using maximal likelihood principle. Variations of this method was introduced by Shuller et al. in 2003 in \cite{schuller2003hmm} and by Lee et al. in 2004 in \cite{lee2004phoneme}. Another common approach is to gather a global statistics over local low-level features computed over the parts of the signal and apply a classification model. Eyben et al. in 2009 \cite{eyben2009openear} and Mower et al. in 2011 \cite{mower2011profiles} used this approach with Support Vector Machine as a classification model. Lee et al. in 2011 in \cite{lee2011decisiontrees} used Decision Trees and Kim et al. in 2013 in \cite{kim2013knn} utilized K Nearest Neighbours instead of SVM. People also tried to adapt popular speech recognition methods to the task of emotion recognition: for more information look at works of Hu et al. in 2007 \cite{hu2007gmm} and Nwe et al. in 2013 in \cite{nwe2013bhattacharyya}.

One of the first deep learning end-to-end approaches was presented by Han et al. in 2014 in their work \cite{han2014elm}. Their idea is to split each utterance into frames and calculate low-level features at the first step. Then authors used densely connected neural network with three hidden layers to transform this sequence of features to the sequence of probability distributions over the target emotion labels. Then these probabilities are aggregated into utterance-level features using simple statistics like maximum, minimum, average, percentiles, etc. After that the Extreme Learning Machine (ELM) \cite{huang2006elm} is trained to classify utterances by emotional state.

In the continuation of the Han et al. work Lee and Tashev presented their paper \cite{lee2015ctc} in 2015. They have used the same idea and approach as Han et al. in  \cite{han2014elm}. The main contribution is that they replaced simple densely-connected network with recurrent neural network (RNN) with Long short-term memory (LSTM) units. Lee and Tashev have also introduced probabilistic approach to learning which is in some points similar to approach presented in current paper. But they continued to use local probabilities aggregation into gloabal feature vector and ELM on top of them.

The main drawbacks of these two approaches are that they are using very simple and naive aggregation functions and ELMs. The latter is actively criticized by the research community last years and Yann LeCun in particular \cite{lecun2015elm}.

This work in its first edition was written in early 2017 \cite{chernykh2017emoctc} and was aimed to get rid of the drawbacks discussed above by applying fully end-to-end pipeline without handcrafted parts in the middle.

After that few purely deep learning and end-to-end approaches based on modern architectures have already arisen. Neumann and Vu in their 2017 paper \cite{neumann2017attentive} used currently popular attentive architecture. Attention is a mechanism that was firstly introduced by Bahdanau et al. in 2015 in \cite{bahdanau2015attention} and now is state-of-the-art in the field of machine translation \cite{vaswani2018attention}. Xia et al. in their 2017 work \cite{xia2017vad} used a slightly different approach based in Deep Belief Networks (DBN) and continuous problem statement in 2D Valence-Arousal space. Each utterance can be assessed in ordinal scale and then embedded into multidimensional space. Regions in this space are associated with different emotions. The task then is to learn how to embed the utterances in this space. One of the most recent and interesting works was presented in 2018 by Lakomkin et al. in \cite{lakomkin2018finetuning}. They suggested to do a transfer learning from usual speech recognition task to the emotion recognition. One might anticipate this method to work well because the speech corpora for speech recognition are far better developed - they are bigger and better annotated. Authors performed a fine-tuning of the DeepSpeech \cite{amodei2016deepspeech} kind of network trained on LibriSpeech \cite{panayotov2015librispeech}.

In spite of existence of few more recent papers on this topic, the quality of the model proposed in this paper is on par with them. At the same time it allows for some extensions like the sequence of emotion labels as an output which other approaches do not support to the best of authors' knowledge.


\section{Data} \label{sec:data}

All experiments are carried out with audio recordings from the Interactive Emotional Dyadic Motion Capture (IEMOCAP) database \cite{busso2008iemocap}. There are also few more emotional speech databases the overview of which can be found in \cite{ververidis2003review, aaac14emodata}. IEMOCAP is chosen because it has one of the most elaborate acquisition methodology, free academic license, long recordings duration and good markup.


\subsection{Database structure} \label{ssec:data_structure}

IEMOCAP \cite{busso2008iemocap} consists of approximately 12 hours of recordings. Audio, video and facial keypoints data was captured during the live sessions. Each session is a sequence of dialogues between man and woman. In total 10 people split into 5 pairs took part in the process. All involved people are professional actors and actresses from Drama Department of University of Southern California \cite{busso2008iemocap}. The recording process took place at the professional cinema studio. Actors seated across each other at "social" distance of 3 meters. It enables more realistic communication.

Before the recording actors were given with the topic of the conversation and the emotional tone in which they should perform. There are two types of dialogues: scripted (actors were given with the text) and improvised.

\begin{figure}[h]
	\begin{subfigure}{.49\textwidth}
		\centering
		\includegraphics[width=1.0\linewidth]{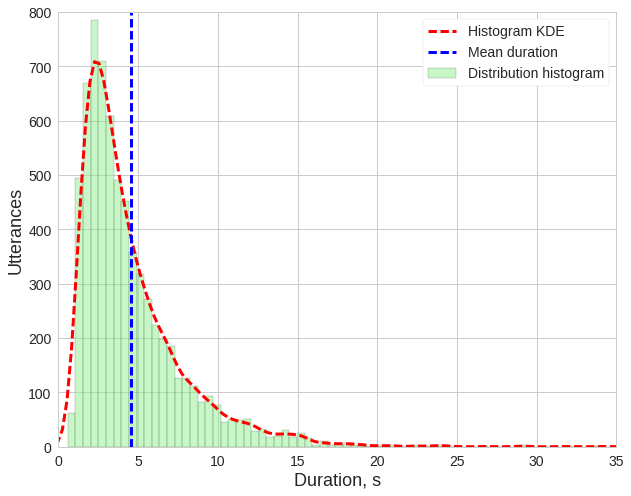}
		\caption{Utterance duration distribution}
		\label{fig:duration_distribution}
	\end{subfigure}
	\begin{subfigure}{.49\textwidth}
		\raggedleft
		\includegraphics[width=1.0\linewidth]{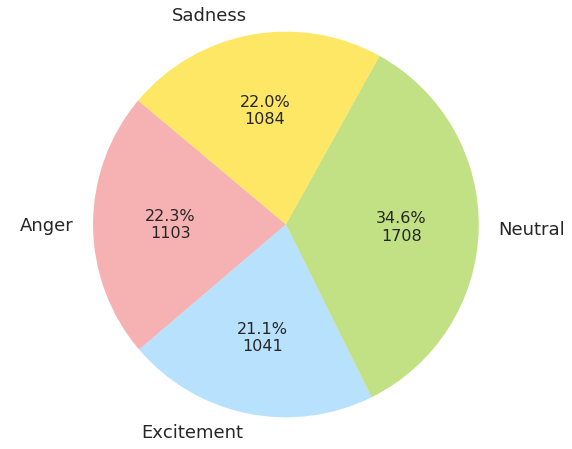}
		\caption{Emotional labels distribution}
		\label{fig:class_distribution}
	\end{subfigure}
	\caption{Data overview}
	\label{fig:data_overview}
\end{figure}

After recording of these conversations authors divided them into utterances with speech (see figure \ref{fig:duration_distribution}).

Note that audio was captured using two microphones. Therefore the recordings contain two channels which correspond to male and female voices. Sometimes they interrupt each other. In these moments the utterances might intersect. This intersection takes about 9\% of all utterances time. It might lead to undesired results because microphones were place relatively near each other and thus inevitably captures both voices.

After the recording assessors (3 or 4) were asked to evaluate each utterance based on both audio and video streams. The evaluation form contained 10 options (neutral, happiness, sadness, anger, surprise, fear, disgust, frustration, excited, other). In this work only only 4 of them are taken for the analysis: anger, excitement, neutral and sadness (as ones of the most common, \cite{picard1995commonemo}). Figure \ref{fig:class_distribution} shows the distribution of considered emotions among the utterances.

\begin{figure}[h]
	\begin{subfigure}{.49\textwidth}
		\centering
		\includegraphics[width=1.0\linewidth]{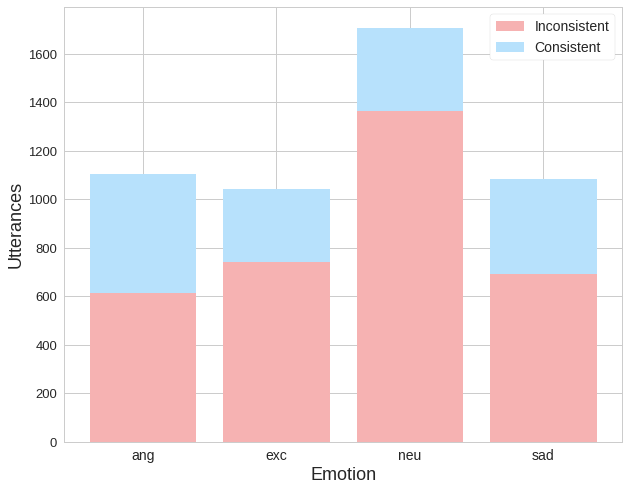}
		\caption{Expert consistency}
		\label{fig:experts_inconsistency}
	\end{subfigure}
	\begin{subfigure}{.49\textwidth}
		\raggedleft
		\includegraphics[width=1.0\linewidth]{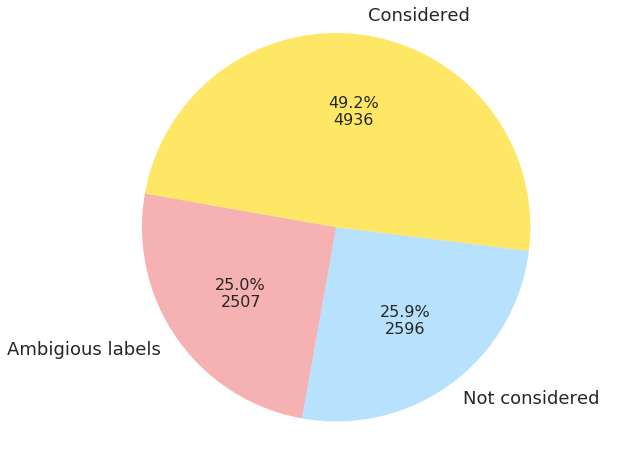}
		\caption{Utterances taken for the work}
		\label{fig:utterances_taken}
	\end{subfigure}
	\caption{Markup details}
	\label{fig:markup_details}
\end{figure}

Emotion is assigned to the utterance if and only if at least half of experts were consistent in their evaluation. About 25\% of the utterances do not satisfy this condition and emotion label was not assigned at all (see figure \ref{fig:utterances_taken}). Moreover, significantly less than a half of remained utterances have consistent assessment from all the experts (figure \ref{fig:experts_inconsistency}). This statistics confirms the statement from section \ref{ssec:problem} that emotion is a subjective notion. Therefore it is reasonable to assume that there is no way to classify emotions accurately even if humans fail to do so.


\subsection{Preprocessing} \label{ssec:preprocessing}

The raw signal has the sample rate of 16 kHz and thus working with it requires enormous computational power. There are technologies (e.g. Google Wavenet \cite{deepmind2016wavenet, deepmind2017wavenet}) that deal with it but for now these algorithms can hardly work online even with Google computational power.

The goal is to reduce the amount of computations down to the acceptable while preserving as much information as possible. Each utterance is divided into intersecting intervals (frames) of 200 milliseconds (overlap by 100 milliseconds). Then acoustic features are calculated over each frame. The resulted sequence of feature vectors represents initial utterance in low dimensional space ans serves as an input to the model.

Authors also experimented with different frame durations from 30 milliseconds to 200 milliseconds. 30 milliseconds roughly correspond to the duration of one phoneme in the normal flow of spoken English. 200 milliseconds is the approximate duration of one word. Experiments do not show significant difference in terms of quality. But computation time rises with the reduction in frame duration due to bigger number of frames. Thus authors decided to stay with 200ms.

Note that labels are presented only for utterances. It means that the task is weakly labelled in a sense that not every frame is labelled.

The key point here is the set of features to calculate. All possible features can be classified into 3 buckets:
\begin{itemize}
    \item Acoustic
    
    They describe the wave properties of a speech. It includes Fourier frequencies, energy-based features, Mel-Frequency Cepstral Coefficients (MFCC) and similar.
    
    \item Prosodic
    
    This type of features measures peculiarities of speech like pauses between words, prosodies and loudness. These speech details depend on a speaker, and use of them in the speaker-free systems is debatable. Therefore they are not used in this work.
    
    \item Linguistic 
    
    These features are based on semantic information contained in speech. Exact transcriptions require a lot of assessor's work. In future it is possible to include speech recognition to the pipeline to use automatically recognized text. But for now authors do not use linguistic features.
\end{itemize}

The current feature extraction algorithm utilizes only acoustic features. PyAudioAnalysis \cite{giannakopoulos2015pyaudioanalysis} library by Giannakopoulos is used. More precisely, 34 features are calculated:
\begin{itemize}
    \item 3 Time-domain: zero crossing rate, energy, entropy of energy
    \item 5 Spectral-domain: spectral centroid, spectral spread, spectral entropy, spectral flux, spectral rolloff
    \item 13 MFCCs
    \item 13 Chroma: 12-dimensional chroma vector, standard deviation of chroma vector
\end{itemize}

In future authors plan to get rid of the handcrafted features and switch to the Convolutional Neural Network (CNN) based feature extraction algorithm.

The final output of the preprocessing step is the sequence of 34-dimensional vectors for each utterance. The length of the sequence depends on the duration of the utterance.


\section{Method} \label{sec:method}

In this paper the Connectionist Temporal Classification (CTC) \cite{graves2006ctc} approach is used to classify speakers by emotional state from the audio recording.

The raw input data is the sound signal which is high-frequency time series. After all the preprocessing steps described in section \ref{ssec:preprocessing} this sound signal is represented as a sequence of multidimensional frame feature vectors. The task is to map this long input sequence into short sequence of emotions which are presented in the recording.

The major difficulty is the significant difference in input and output sequences lengths. The input sequence length might be about 100 which is about 10 seconds with the chosen preprocessing settings. Output sequence length is usually no more than 2-4. Two orders of magnitude difference. In this case usual solutions such as padding of output sequence of bucketing (which is used in Google Neural Machine Translation \cite{google2016nmt}) can hardly be applied.

CTC addresses this problem in an essential way by utilizing three main concepts:
\begin{itemize}
    \item Introduce additional NULL label which corresponds to the absence of any other label and extends the initial labels set.
    \item Bijective sequence-to-sequence learning, i.e., one-to-one mapping from sequence of frame features to the sequence of extended labels.
    \item Collapse resulting sequence w.r.t. duplicates of labels and introduced extra label.
\end{itemize}

In case of emotion recognition these features are inherently implied by the essence of the task. On the one hand one utterance may contain several different emotions but on the other hand there might be considerable parts of recording without any sign of emotions.

Thus there are strong reasons to believe that one can benefit from usage of Connectionist Temporal Classification approach in this problem.


\subsection{Notation} \label{ssec:notation}

Let $E = \{0 \ldots k-1\}$ be the set of labels and $L = E \cup \{\text{NULL}\}$ --- extended label set.

Assume that $\mathcal{D} = \{(\text{X}_i, \mathbf{z}_i)\}_{i=1}^{n}$ is the dataset where $\mathbf{z}_i \in \mathcal{Z} =  E^*$ is the true sequence of labels and $\text{X}_i \in \mathcal{X} = \left( \mathbb{R}^f \right)^*$ --- corresponding $f$-dimensional feature sequence. It is worth to mention that the lengths of these sequences $|\mathbf{z}_i| = U_i$ and $|\text{X}_i| = T_i$ may not be the same in general case, the only condition is that $U_i \leq T_i$. 

Next let's introduce the set of decision functions or models $\mathcal{F} = \{ f : \mathcal{X} \mapsto \mathcal{Z} \}$ in which the best model is to be found. In case of neural network with the fixed architecture it is essential to associate the set of functions $\mathcal{F}$ with the network weights space $\mathcal{W}$ and thus function $f$ and vector of weights $\mathbf{w}$ are interchangeable.

Having the set of functions one need to know how to choose the best. For that purpose probabilistic approach and maximal likelihood training is used (one can learn more in \cite{bishop06prml}). Assume that the model $f$ can also calculate the probability measure $\text{p}$ of any sequence being its output. Then one wants the likelihood of the dataset $\mathcal{D}$ to be as high as possible:
\begin{equation*}
    \prod_{i=1}^{|\mathcal{D}|}\text{p}\left(\mathbf{z}_i|\text{X}_i\right) \rightarrow \max.
\end{equation*}
The optimal model then can be found as:
\begin{equation*}
    \hat{f} = \argmax\limits_{f \in \mathcal{F}} \sum_{i=1}^{|\mathcal{D}|} \log\text{p}\left(\mathbf{z}_i|\text{X}_i\right) = \argmin\limits_{\mathbf{w} \in \mathcal{W}} Q\left( \mathbf{w}, \mathcal{D}\right).
\end{equation*}
This method can be seen from the angle of loss functions and Empirical Risk Minimizer (see \cite{mohri12foundationsml})

In case of neural network models the optimization is usually carried out with gradient descent type algorithms.


\subsection{CTC approach} \label{ssec:ctc}

CTC is the one of the sequence-to-sequence prediction methods that deals with the different lengths of the input and output sequences. The main advantage of CTC is that it chooses the most probable label sequence (labeling) regarding the various ways of aligning it with the initial sequence. The probability of the particular labeling is added up from the probabilities of every its alignment.

\begin{figure}[h]
    \centering
    \includegraphics[width=1.0\linewidth]{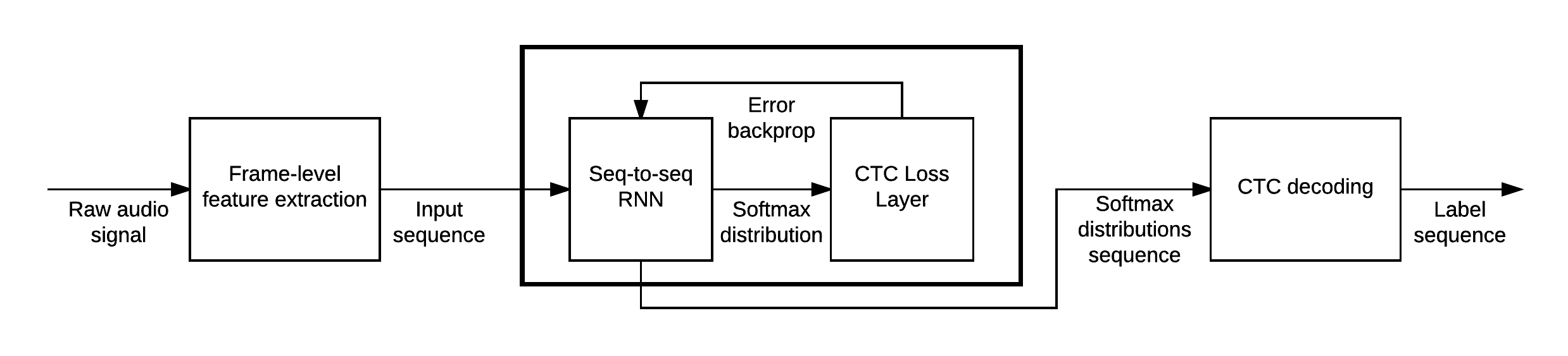}
    \caption{CTC pipeline} \label{fig:ctc_diagram}
\end{figure}

In the figure \ref{fig:ctc_diagram} the pipeline of the CTC method is depicted.

Recurrent neural network (RNN) with fixed architecture (see details in section \ref{ssec:ctc_exp}) is chosen as a space of classifiers $\mathcal{F}$. The only requirement for the structure is to output the sequence of the same length as it takes as an input.

Think of RNN as a mapping from the input space $\mathcal{X}$ to the sequence of probability distributions over the extended label set $L$:
\begin{equation*}
    \text{Y} = f\left(\text{X}\right) \in [0; 1]^{(k + 1) \times T},
\end{equation*}
where $y_c^t$ is the output of the softmax layer and represents the estimation of the probability of observing class $c$ at the timestep $t$.

For every input $\text{X}$ let's define the path $\bm{\pi}$ --- it is an arbitrary sequence from $L^*$ with the length of $T$. Then the conditional probability of the path is
\begin{equation*}
    \text{p}(\bm{\pi}|\text{X}) = \prod_{t=1}^{T} y_{\pi_t}^t.
\end{equation*}
The problem is that the path can contain $\text{NULL}$ class which is unacceptable in the final output. First of all one needs to get rid of the $\text{NULLs}$. For that purpose mapping $M : L^T \mapsto E^{\leq T}$ is introduced. It basically consists of two steps:

\begin{enumerate}
    \item Delete all consequent repeated labels
    \item Delete all $\text{NULLs}$
\end{enumerate}

Consider the following example: $M \left( -aa-b-b--ccc \right) = M \left( abb---bc- \right) = abbc$. Notice that $M$ is the surjective mapping. By means of it the paths are transformed into labelings. To compute the probability of the labeling one needs to sum up probabilities of all paths that wrap into this particular labeling:
\begin{equation*}
    \text{p}(\mathbf{l} | \text{X}) = \sum_{\bm{\pi} \in M^{-1}(\mathbf{l})} \text{p}(\bm{\pi} | \text{X}).
\end{equation*}
The direct calculation of $\text{p}(\mathbf{l}|\text{X})$ requires summation over all corresponding paths which is exhaustive task. There are $(k+1)^{T}$ possible paths. Graves et al. \cite{graves2006ctc} derived a new efficient forward-backward dynamic programming algorithm for that. The initial idea was taken from HMM decoding algorithm introduced by Rabiner \cite{rabiner1989hmm}.

Finally, the objective function is
\begin{equation*}
    Q(\mathbf{w}, \mathcal{D}) = 
    -\sum_{i=1}^{|\mathcal{D}|}\log \text{p}(\mathbf{z}_i|\text{X}_i) = 
    -\sum_{i=1}^{|\mathcal{D}|}\sum_{\bm{\pi} \in M^{-1}(\mathbf{z}_i)}\log \text{p}(\bm{\pi}|\text{X}_i).
\end{equation*}
Neural network here plays a role of probability measure $\text{p}$ evaluator and the more it trains the more accurate probability estimations it gives. To enable the neural network training with the standard gradient-based methods Graves et al. \cite{graves2006ctc} suggested differentiation technique naturally embedded into dynamic programming algorithm.

The final model chooses the labeling with the highest probability:
\begin{equation*}
    \mathbf{h}(\text{X}) = \argmax_{\mathbf{l} \in E^{\leq T}}\text{p}(\mathbf{l}|\text{X})
\end{equation*}
However one has exponential number of labelings and thus the task of accurate probability computation is intractable. There are two main heuristics for tackling this problem:

\begin{enumerate}
    \item Best path search 
    
    It approximates the most probable labeling with the wrapped version (after $M$ transformation) of the most probable path.
    
    \item Beam search
    
    It keeps track of the fixed length prefix to choose the most probable label at each step. Best path search is a special case of beam search where the beam width equals to 1.
\end{enumerate}

Both heuristics are tested during the experiments.


\section{Experiments} \label{sec:experiments}

In the series of experiments authors investigate proposed approach and compare it to the different baselines for emotion recognition. All the code can be found in the github repository \cite{chernykh18github}.

One of the main obstacles with the speech recognition task in practice is that it is usually weakly supervised (as described in section \ref{ssec:preprocessing}). Here it means that there are a lot of frames in the utterance but only one emotional label. At the same time it is obvious that for sufficiently long periods of speech not all the frames contain emotions. CTC loss function suggests one of the ways to overcome this issue.

Authors choose two more methods and provide a comparison between them and CTC in the same setting. The algorithms are described at section \ref{ssec:baselines} while the results are reported at section \ref{ssec:comparison}.

In all the methods and algorithms discussed below the frame features are calculated as described in section \ref{ssec:preprocessing}.

Please also note, that in IEMOCAP database each utterance has only one emotion. Therefore in CTC approach the length of all the real output sequence equals to one $U_i = |\mathbf{z}_i| = 1$. Thus one can consider the output sequence of emotion labels as one emotion assigned to the utterance and vectors $\mathbf{z}_i$, $\mathbf{h}(\text{X}_i)$ as scalars $z_i$, $h_i$.


\subsection{Metrics} \label{ssec:metrics}

First of all, one need to decide on the evaluation criteria. In this work authors follow the suggestion from Lee et al. \cite{lee2015ctc} and uses two main metrics to evaluate and compare the models:

\begin{itemize}
    \item Overall (weighted) accuracy
    \begin{equation*}
        \frac{1}{n}\sum_{i=1}^{n}\left[z_i=h_i\right]
    \end{equation*}
    It is a usual accuracy which is calculated as a fraction of correct answers over all examples.
    \item Mean class (unweighted) accuracy
    \begin{equation*}
        \frac{1}{c}\sum_{c=1}^{k}\frac{\sum\limits_{i=1}^{n}\left[z_i=h_i\right]\cdot\left[z_i=c\right]}{\sum\limits_{i=1}^{n}\left[z_i=c\right]}
    \end{equation*}
    The idea is to take accuracy only inside one class and then average these values across all classes.
\end{itemize}

In both formulas above the square brackets denote indicator function.

Overall accuracy is the standard metric which is common to use and thus easy to compare with the results from other papers. But it has one major drawback. It does not account for the class imbalance. While in the case of IEMOCAP dataset, e.g., neutral class is approximately 1.7x times bigger than excitement. Therefore authors introduce mean class accuracy which taked into account the differences in class sizes and get rid of the imbalance influence on the metric value.


\subsection{Baselines} \label{ssec:baselines}

In this subsection one can find the description and the performance report of the baselines algorithms.


\subsubsection{Framewise} \label{sssec:framewise_exp}

The core idea of this method is to classify each frame separately. Remember that the task is weakly supervised the following workflow is chosen:
\begin{itemize}
\item Take two loudest frames from each utterance. Loudness in this context equals to the spectral power
\item Assign these frames with the emotion of the utterance
\item Train the frame classification model 
\item Label all frames in all utterances using fitted model
\item Classify utterances based on the obtained frame-level labels
\end{itemize}

The naive assumption here is that the whole utterance can be represented by 2 loudest frames.

Random Forest Classifier \cite{sklearn2011} is used as a classification model. 
To assign emotion to the utterance majority voting is applied to the emotion labels of the frames. More detailed description of the algorithm, hyperparameters setting and code might be found in the github repository \cite{chernykh18github}.

\begin{figure}[h]
    \captionsetup{justification=centering}
    \centering
    \includegraphics[width=1.0\linewidth]{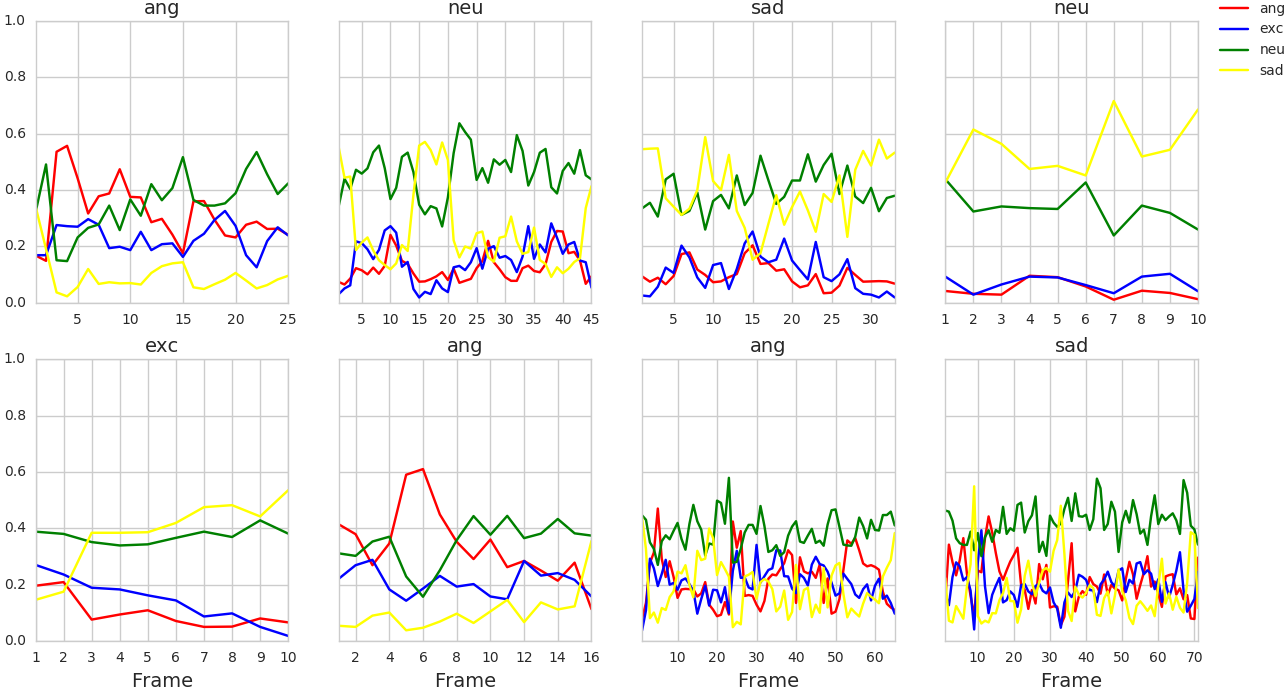}
    \caption{Framewise classification model. The title of each plot is the real emotion of the utterance. Each emotion is depicted with the color, x-axis shows the number of frame, y-axis gives the probability of classifying the frame with the emotion.}
    \label{fig:framewise_rfc}
\end{figure}

In the figure \ref{fig:framewise_rfc} there are the results of this method for randomly chosen validation set utterances. One can observe that for short utterances it works fine but with longer utterances it becomes sawtooth and unstable.

For the methodology and results of the overall comparison with other methods please see section \ref{ssec:comparison} and table \ref{tab:accuracy}.


\subsubsection{One-label} \label{sssec:onelabel_exp}

One-label approach implies that every utterance has only one emotional label notwithstanding its length. In other words sequence-to-label learning paradigm is used here in contrast with sequence-to-sequence learning in CTC.

The important detail is that all major modern deep learning frameworks (like TensorFlow, Keras, PyTorch, etc.) can group data into batches. Batch is in fact a multidimensional tensor. Mini-batch gradient descent and its modifications is the de facto standard method of training for neural networks. But the peculiarity here is that only the tensors of the same dimensions can be packed into the batch. After the preprocessing steps described in section \ref{ssec:preprocessing} the input data is the sequences of the same dimension (34) but of the different length which depends on the duration of the utterance. Thus it is impossible to pack them into batch and train a network efficiently.

There are couple of solutions to this problem, e.g., padding or bucketing \cite{google2016nmt}. Here authors use padding. The idea is to make all the sequences of the same length. For that short sequences are appended with zeros and long sequences are truncated to the unified length. In this work the unified length equals to 78 which is approximately the 90\%-percentile of all sequences lengths. After that step the training can be done efficiently using mini-batch approaches. Authors used Adam \cite{kingma2014adam} optimizer for the training.

One-label approach also requires the definition of the network architecture. Authors decided to use same architecture for all of the approaches to be able to fairly compare them. One-label architecture is depicted in the figure \ref{fig:cce_blstm} of Appendix A. It contains stacked Bidirectional LSTM units and dense classification layers on top of them. Categorical cross-entropy loss function is used. For more detailed description of the network structure and training procedure see figure \ref{fig:cce_blstm} in Appendix A and code in \cite{chernykh18github}.

The methodology and results of the overall comparison with other methods are described section \ref{ssec:comparison} and table \ref{tab:accuracy}.


\subsection{CTC} \label{ssec:ctc_exp}

\begin{figure}[h]
    \captionsetup{justification=centering}
	\begin{subfigure}{.49\textwidth}
		\centering
		\includegraphics[width=1.0\linewidth]{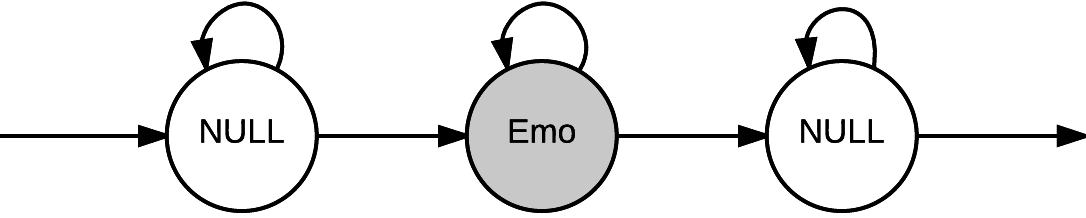}
	\end{subfigure}
	\begin{subfigure}{.49\textwidth}
		\centering
		\includegraphics[width=1.0\linewidth]{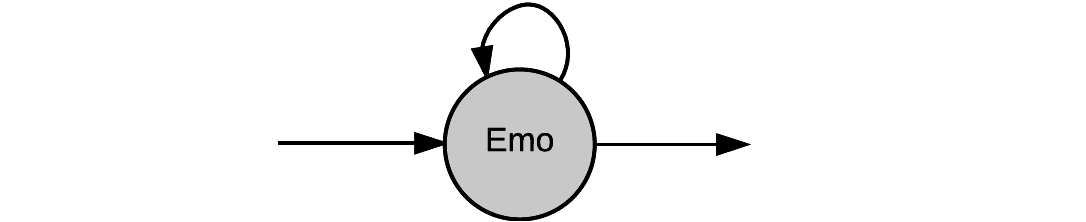}
	\end{subfigure}
	\vfill
	\begin{subfigure}{.49\textwidth}
		\centering
		\includegraphics[width=1.0\linewidth]{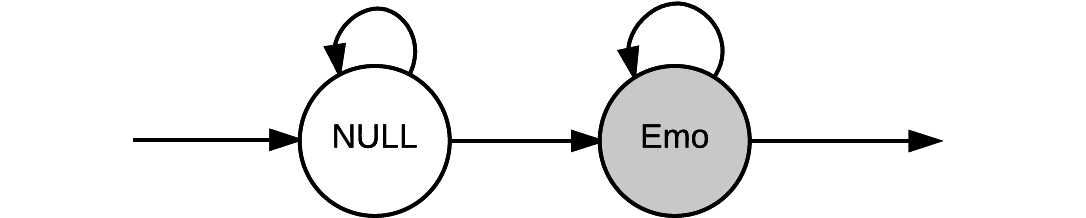}
	\end{subfigure}
	\begin{subfigure}{.49\textwidth}
		\centering
		\includegraphics[width=1.0\linewidth]{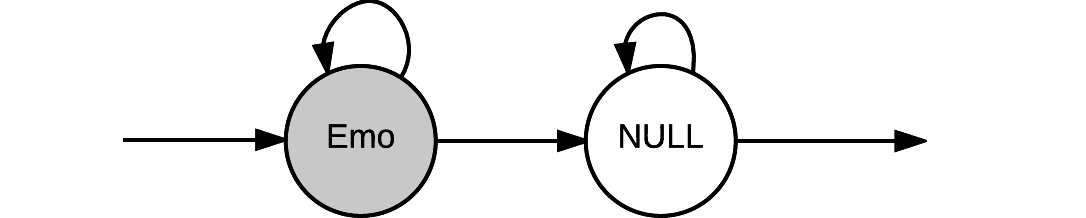}
	\end{subfigure}
	\caption{Valid sequences of labels. "Emo" label in all schemes represents exactly one emotion. It can be one and only one of the 4 emotion discussed in \ref{ssec:data_structure}: anger, excitement, neutral and sadness}
	\label{fig:valid_seqs}
\end{figure}

Although CTC approach can inherently account for more than one label in the utterance, the design of the IEMOCAP database implies only one emotion per utterance (see sections \ref{ssec:preprocessing} and \ref{sec:experiments}). Consequently there are four valid types of label sequences from $L^*$ which can be generated by the network (see figure \ref{fig:valid_seqs}).

Each type of the sequence is later collapsed by the $M$ transformation during CTC decoding step (see section \ref{ssec:ctc}). Note that all 4 valid sequence types are collapsed into one "Emo" label.

\begin{figure}[h]
    \captionsetup{justification=centering}
	\centering
	\includegraphics[width=1.0\linewidth]{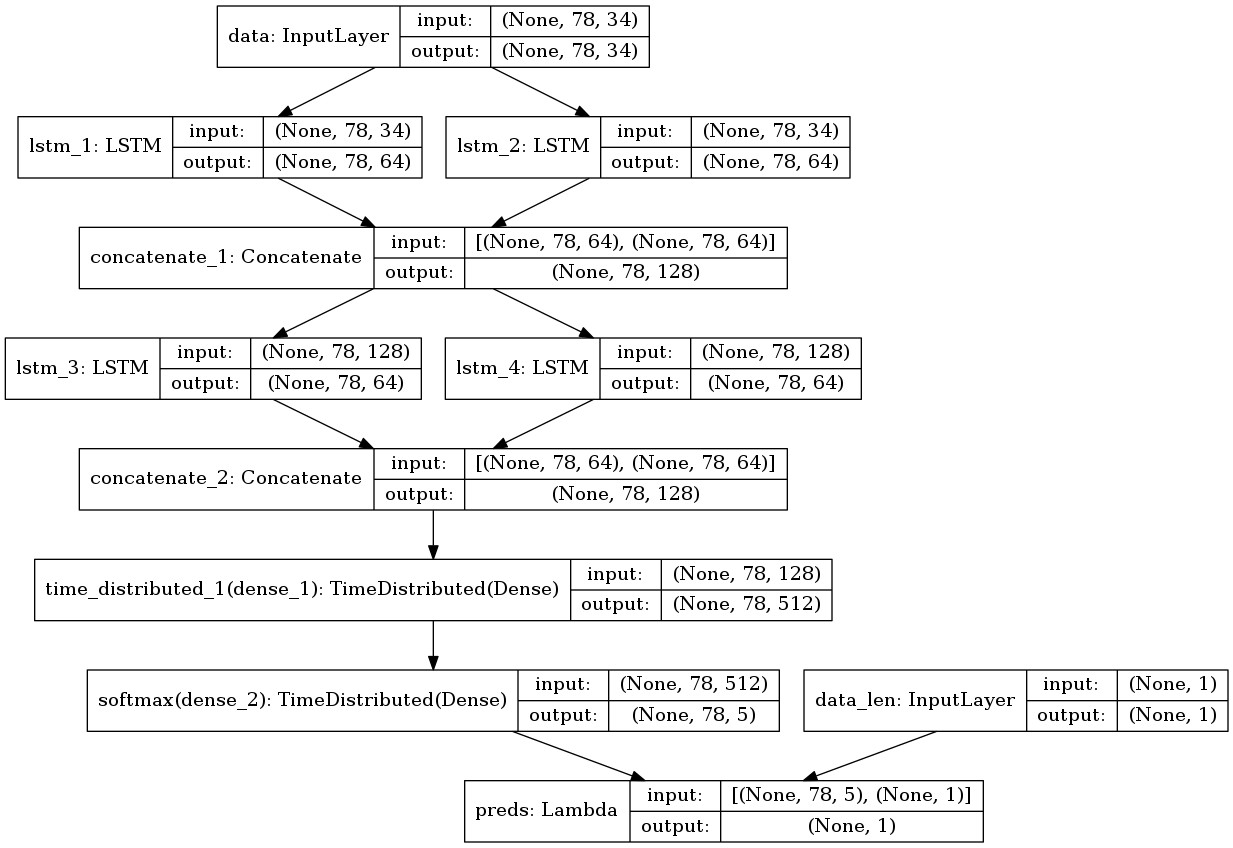}
	\caption{BLSTM network architecture for CTC approach. $\text{lstm\_1}$ and $\text{lstm\_3}$ layers process the input sequences in the forward order while $\text{lstm\_2}$ and $\text{lstm\_4}$ do it in the backward order. After processing the sequence in the backward order the output of $\text{lstm\_2}$ and $\text{lstm\_4}$ is reversed one more time to be in the forward order. After that outputs of $\text{lstm\_1}$, $\text{lstm\_2}$ and $\text{lstm\_3}$, $\text{lstm\_4}$ are stacked as shown. $\text{TimeDistributed}$ Keras wrapper allows to apply one and the same dense layer to each element of the input sequence. Last $\text{Lambda}$ layer allows to perform CTC decoding step. Additional input layer $\text{data\_len}$ contains the real length of the initial input sequence (without padding) which allows for more precise decoding.}
	\label{fig:ctc_blstm}
\end{figure}

When applying the CTC approach one faces the same problem with different input sequence lengths as one saw in One-label approach in section \ref{sssec:onelabel_exp}. The solution here is the same. Input sequences are padded or truncated to the length of 78. The only difference is that one keeps track of the initial sequence length to decode the resulting output sequence even better by not taking into account padded places (see figure \ref{fig:ctc_blstm} and code \cite{chernykh18github} for more details).

CTC approach requires the neural network architecture. As it is mentioned in section \ref{sssec:onelabel_exp} authors decided to use same architecture for all of the approaches to be able to fairly compare them. CTC architecture is shown in the figure \ref{fig:ctc_blstm} of Appendix A. It contains stacked Bidirectional LSTM units and dense classification layers on top of them. CTC loss function is used. For more detailed description of the network structure and training procedure see figure \ref{fig:ctc_blstm} in Appendix A and code in \cite{chernykh18github}.

The methodology and results of the overall comparison with other methods are described section \ref{ssec:comparison} and table \ref{tab:accuracy}.


\subsection{Comparison} \label{ssec:comparison}

In this section we provide a comparison between all three approaches described above in sections \ref{sssec:framewise_exp}, \ref{sssec:onelabel_exp}, \ref{ssec:ctc_exp}.

Each method is tested using grouped cross-validation approach. In usual k-fold cross-validation approach the dataset is randomly split into into k disjoint folds. At each of k steps the the $\text{k}^{\text{th}}$ fold is used as a test set and all other folds are used as a training set.

Grouped cross-validation assumes that each data sample has an additional label. This label shows the group of the sample. Group in this context might be any kind of common property that samples share. In this work the group is a speaker. It means that the group labels contains all samples that were spoken by one person (and only them). Grouped cross-validation splits the data in such a way that samples from the one group can not be in both training and test sets simultaneously.

Grouped cross-validation technique allows to ensure that the model quality is measured in speaker independent way. It means that the model is not overfitted to the manner of particular speakers presented in the training set.

IEMOCAP dataset contains 10 speakers which were recorded by pairs. Each speaker has roughly the same number of utterances. If one was to split the data into groups according to the speaker then one would get only 10\% of data for the test. That might be to unstable. Thus authors decided to form groups not by speakers exactly but by pair of speakers that were recorded simultaneously. In that way 20\% of data is split for the test which is more stable.

The results of 5-fold grouped cross-validation averaged across folds are shown in the table \ref{tab:accuracy}.

\begin{table}[h]
	\centering
	\caption{Methods comparison}
	\label{tab:accuracy}
	\vspace{2mm}
	\begin{tabular}{@{}ccc@{}}
		\toprule
		\textbf{Method} & \textbf{Overall accuracy}  & \textbf{Mean class accuracy} \\ \midrule
		Dummy           & 35\%                       & 25\%                         \\
		Framewise       & 45\%                       & 41\%                         \\
		One-label       & 51\%                       & 49\%                         \\
		CTC             & 54\%                       & 54\%                         \\ \midrule
		Human           & 69\%                       & 70\%                         \\ \bottomrule
	\end{tabular}
\end{table}

First row with "Dummy" method corresponds to the naive classification model which always answers with label of the largest training class. In IEMOCAP case it is neutral class. "Framewise" and "One-label" rows represent the described baseline models. "CTC" shows the model investigated in this paper. As one can notice CTC performs slightly better than One-label approach and much better than Framewise and Dummy.

The last line in this table shows the human performance at the same task. Authors conducted the series of experiments to measure it. This process is described in more details in section \ref{ssec:human_performance}.


\subsection{Error structure} \label{ssec:error_structure}

Observing the quality of the CTC model in section \ref{ssec:comparison} authors also decided to further investigate it. Graves et al. in \cite{graves2006ctc} reports huge gap in quality over the classical models. Here the gain is about 3-5\%. For that reason the error structure is studied.

\begin{figure}[h]
	\begin{subfigure}{.49\textwidth}
		\centering
		\includegraphics[width=1.0\linewidth]{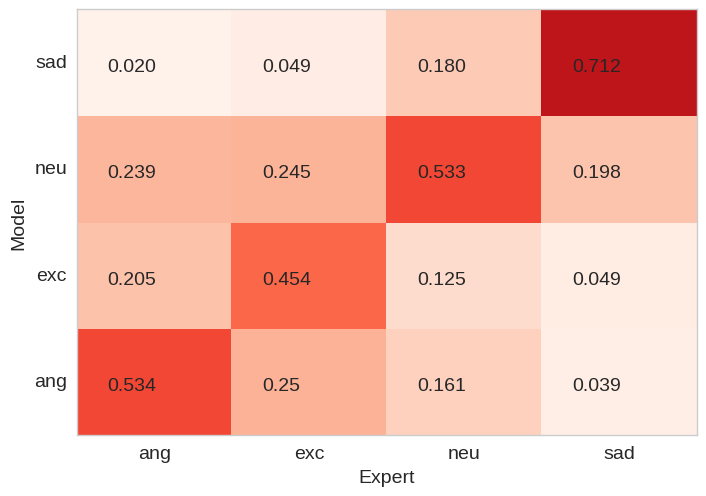}
		\caption{Confusion matrix}
		\label{fig:ctc_confusion}
	\end{subfigure}
	\begin{subfigure}{.49\textwidth}
		\centering
		\includegraphics[width=1.0\linewidth]{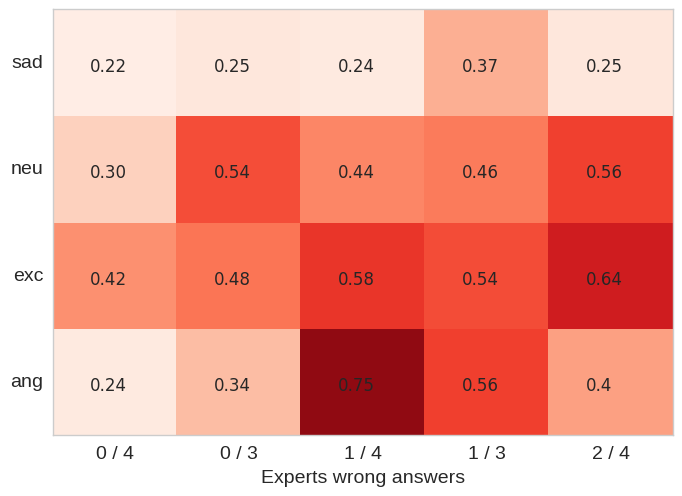}
		\caption{Misclassification rates}
		\label{fig:ctc_misclassification}
	\end{subfigure}
	\caption{CTC BLSTM error structure}
	\label{fig:ctc_error}
\end{figure}

First of all, let's look at predictions distribution in comparison with real expert labels. This is done by means of confusion matrix shown in the figure \ref{fig:ctc_confusion}. Busso et al. in \cite{busso2008iemocap} mention that audio signal plays the main role in sadness recognition while angry and excitement are better detected via video signal which accompanied audio during the assessors work on IEMOCAP. This hypothesis seems to be true in relation to CTC model. Sadness recognition percentage is much higher than the others.

In section \ref{ssec:data_structure} authors have already described that expert answers are not fully consistent sometimes (see figure \ref{fig:experts_inconsistency}). It allows to speak about the reliability of the label. Figure \ref{fig:ctc_misclassification} shows how the model quality depends on the expert confidence degree. On the x-axis one can see the number of experts whose answer differs from the final emotion assigned to the utterance. y-axis shows the emotion label. In each cell of a table there is a model error percentage when classifying corresponding emotion at corresponding confidence level. The more red the cell is the the bigger the error is.

In fact this matrix gives an interesting piece of information. If one takes in account only those utterances in which experts were consistent then one gets approximately 65\% accuracy. It sounds more promising than 54\%.

Going further, authors investigate the wrong predictions themselves and not only their distribution. In inconsistent samples some experts give answers that are not the same as the final emotion assigned to the utterances. These answers can be arbitrary emotion from the full IEMOCAP list. Here authors filter only four considered emotions from all the wrong answers.

In the first row of the table \ref{tab:residual_accuracy} there is the percentage of inconsistent answers from utterances labeled as the header name which falls into considered four emotions. For example, 17\% in column "Anger" means the following: utterances finally labeled as angry have some inconsistent expert answers; 17\% of these answers have labels from the set of considered 4 emotions.

In the second row there is the percentage of model answers that coincide with the inconsistent answer of expert in this case. Note that there can not be more than one inconsistent answer because otherwise half of the experts would be inconsistent and utterance should not be included into the dataset at all.

\begin{table}[h]
	\centering
	\caption{Residual accuracy}
	\label{tab:residual_accuracy}
	\vspace{2mm}
	\begin{tabular}{@{}ccccc@{}}
		\toprule
		                 & \textbf{Anger} & \textbf{Excitement} & \textbf{Neutral} & \textbf{Sadness} \\ \midrule
		Considered ratio & 17\%           & 22\%                & 36\%             & 39\% \\
		Model accuracy   & 51\%           & 73\%                & 71\%             & 74\% \\
		\bottomrule
	\end{tabular}
\end{table}

In other words, table \ref{tab:residual_accuracy} shows how frequently the errors of our model coincide with the human divergence in emotion assessment. If the errors of the model were random then second row of the table would contain approximately 33\% at each cell. In the case of the CTC model this percentage is much higher. It means that the models make the mistakes which are similar to human mistakes. This topic is further discussed in the section \ref{ssec:human_performance}.


\subsection{Human performance} \label{ssec:human_performance}

Observing the inconsistency of experts and other problems of the markup described in the sections \ref{ssec:error_structure} and \ref{sec:data} authors come with the idea to see how humans perform at this task.

This question was previously arisen in the papers. As authors have already described in the section \ref{ssec:problem}, Altrov et al. did the same work in \cite{altrov2015humanemotions}. They used almost the same 4 classes (joy, anger, sadness, neutral) thus the results might be comparable. Native language speakers scored about 69\% mean class accuracy. All other people perform 10-15\% worse.

\begin{figure}[h]
	\begin{subfigure}{.49\textwidth}
		\centering
		\includegraphics[width=1.0\linewidth]{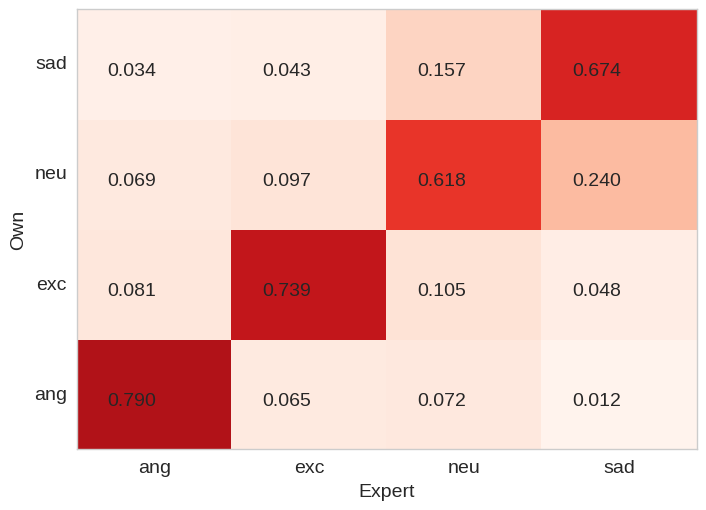}
		\caption{Human error structure}
		\label{fig:human_confusion}
	\end{subfigure}
	\begin{subfigure}{.49\textwidth}
		\centering
		\includegraphics[width=0.8\linewidth]{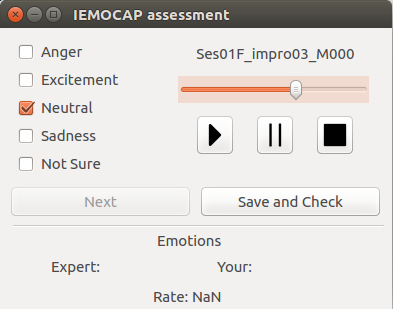}
		\vspace*{5mm}
		\caption{Labelling interface}
		\label{fig:interface}
	\end{subfigure}
	\caption{Human labeling}
	\label{fig:markup_analysis}
\end{figure}

In this work a simple interface (fig. \ref{fig:interface}) for relabelling speech corpus was developed. The idea is to see how well humans can solve this classification task. One can consider that as a humanized machine learning model.

Five people were involved in the experiment. All of them were authors' lab colleagues (not professional actors or psychologist) and their native language is Russian. Each of them was asked to assess the random subset of the utterances. There is a possibility to see the correct answer after one gives own answer. This allows for positive feedback loop and kind of "model training" in terms of humanized machine learning model. During the experiment a small fraction of the utterances (2 from each emotion, 8 in total) was excluded from the main dataset. These utterances were given to the assessors prior to the main experiment as a kind of training examples. Through these mechanism assessors were able to get used to the system, way how actors talk, tune the volume level and other parameters. Answers at these preliminary stage were not included in the final statistics. Finally, each utterance was assessed by at least 2 assessors.

In the figure \ref{fig:human_confusion} one can see the results of the experiment taken.

Both overall accuracy and mean class accuracy are about 70\% (see table \ref{tab:accuracy}). These numbers confirm the idea that the emotion is the subjective notion and it is hardly probable for any model to achieve even this 70\%. In this light the model error structure investigated in the section \ref{ssec:error_structure} becomes crucial because human errors are not random. Humans make mistakes in the cases where the emotion is indeed unclear. For example, it is hard to confuse angry and sadness, but it is easy to do so for excitement and happiness.

It leads to the conclusion that to be able to see the real quality of the model one should look not only at the accuracy numbers but also at the error structure. It should be reasonable and resembles human structure. In case both criteria are satisfied (high enough accuracy and reasonable error structure) one can say that the model is good. Error structure analysis for CTC model which is carried out in section \ref{ssec:error_structure} satisfies both criteria and thus the investigated CTC model can be considered to work well.


\section{Conclusion}

In this paper authors propose a novel algorithm for emotion recognition from audio based on Connectionist Temporal Classification approach. There are two main advantages of the suggested method:

\begin{itemize}
\item It takes into account that even the emotional utterance might contain parts where there is no emotions
\item It can predict the sequence of emotions for one utterance
\end{itemize}

Conducted experiments lead to the results are comparable with the state-of-the-art in this field. Authors provide an in-depth analysis of the models answers and errors. Moving further, the human performance on this task is measured to be able to understand the possible limits of the model improvements. The initial suggestion that emotion is a subjective notion is approved and it turns out that the gap between human and proposed model is not so big. Moreover, the error structure for the humans and the model is similar which becomes one more argument in favor of the model.

Authors have few plans on the future development of the current work. One way is to get rid of the handcrafted MFCC feature extraction and switch to the learnable methods like Convolutional Neural Networks. Another way is to apply domain adaptation techniques and transfer the knowledge from the speech recognition methods to the the emotion detection using pretraining and fine-tuning.

\bibliographystyle{unsrt}
\bibliography{CTC}

\newpage
\section*{Appendix A}
\begin{figure}[h]
    \captionsetup{justification=centering}
	\centering
	\includegraphics[width=1.0\linewidth]{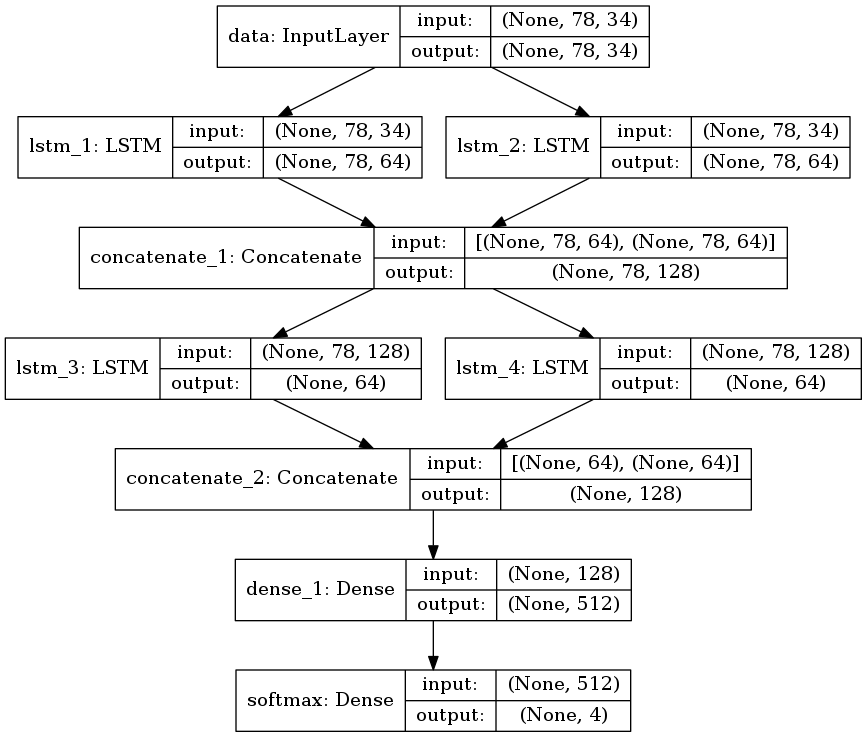}
	\caption{BLSTM network architecture for one-label approach. $\text{lstm\_1}$ and $\text{lstm\_3}$ layers process the input sequences in the forward order while $\text{lstm\_2}$ and $\text{lstm\_4}$ do it in the backward order. After processing the sequence in the backward order the output of $\text{lstm\_2}$ and $\text{lstm\_4}$ is reversed one more time to be in the forward order. After that outputs of $\text{lstm\_1}$, $\text{lstm\_2}$ and $\text{lstm\_3}$, $\text{lstm\_4}$ are stacked as shown. Note that here (in contrast with CTC architecture in figure \ref{fig:ctc_blstm}) last LSTM layers $\text{lstm\_3}$ and $\text{lstm\_4}$ output only the last state and not the whole sequence. Thus one does not need $\text{TimeDistributed}$ Keras wrapper and can go with simple $\text{Dense}$ layer. Last $\text{softmax}$ layer has 4 output units because the chosen subset of IEMOCAP dataset has 4 emotions (see section \ref{ssec:data_structure}). Note also that the real length of the initial input sequence (without padding) is not taken into account in this approach.}
	\label{fig:cce_blstm}
\end{figure}

\end{document}